%% file: main.tex
\documentclass{article}

\usepackage{microtype}
\usepackage{graphicx}
\usepackage{caption}
\usepackage{subcaption}
\usepackage{multirow}
\usepackage{booktabs} 

\usepackage{hyperref}

\usepackage{bm}
\usepackage{xspace}

\usepackage[accepted]{icml2021}

\icmltitlerunning{MAM: Masked Acoustic Modeling for End-to-End Speech-to-Text Translation}

\usepackage{dsfont}
\usepackage{amsmath}
\usepackage{amssymb}
\usepackage{amsthm}

\usepackage{makecell}
\usepackage{multirow}

\usepackage{titlesec}
\titleformat*{\subsubsection}{\bfseries}
\definecolor{tgreen}{rgb}{0,0.7,0.3}
\input{defs}

\begin{document}

\twocolumn[
\icmltitle{MAM: Masked Acoustic Modeling for End-to-End Speech-to-Text Translation}



\icmlsetsymbol{equal}{*}

\begin{icmlauthorlist}
\icmlauthor{Junkun Chen}{equal,to,goo}
\icmlauthor{Mingbo Ma}{equal,to}
\icmlauthor{Renjie Zheng}{to}
\icmlauthor{Kaibo Liu}{to}
\icmlauthor{Liang Huang}{to,goo}
\end{icmlauthorlist}
\icmlaffiliation{to}{Baidu Research, Sunnyvale, CA, USA;}
\icmlaffiliation{goo}{Oregon State University, Corvallis, OR, USA}
\icmlcorrespondingauthor{Mingbo Ma}{cosmmb@gmail.com}
\icmlkeywords{Machine Learning, ICML}
\vskip 0.3in
]


\printAffiliations{\icmlEqualContribution}

\begin{abstract}
End-to-end Speech-to-text Translation (E2E-ST), which directly translates 
source language speech to target language text,
is widely useful in practice, 
but traditional cascaded approaches (ASR+MT) often suffer from error propagation in the pipeline.
On the other hand,
existing end-to-end solutions heavily depend on the source language transcriptions
for pre-training or multi-task training with Automatic Speech Recognition (ASR).
We instead propose a simple technique to 
learn a robust speech encoder in a self-supervised fashion 
only on the speech side, 
which can utilize speech data without transcription.
This technique termed Masked Acoustic Modeling (MAM),
not only provides an alternative solution to improving E2E-ST,
but also can perform pre-training 
on 
any acoustic signals (including non-speech ones) without annotation.
We conduct our experiments over 8 different translation directions.
In the setting without using any transcriptions,
our technique achieves an average improvement of 
+1.1 BLEU,
and +2.3 BLEU with MAM pre-training. 
Pre-training of MAM with arbitrary acoustic 
signals also has an average improvement
with +1.6 BLEU for those languages.
Compared with ASR multi-task learning solution, which replies on transcription during training,
our pre-trained MAM model, which does not use transcription, achieves similar accuracy. 
\vspace{-5pt}
\end{abstract}
\vspace{-10pt}

\section{Introduction}
\input{intro}

\section{Preliminaries: ASR and ST}
\input{prelim}

\section{Masked Acoustic Modeling}
\input{methods}

\section{Experiments}
\input{exp}


\section{Related Work}
\input{related}

\vspace{-5pt}
\section{Conclusions}
We have presented a novel acoustic modeling framework MAM in this paper.
MAM not only can be used as an extra component during training time,
but also can be used as a separate pre-training framework with
arbitrary acoustic signal.
We demonstrate the effectiveness of MAM with multiple 
different experiment settings in 8 languages.
Especially, for the first time,
we show that pre-training with arbitrary acoustic data with MAM  
boosts the performance of speech translation.





\bibliography{main}
\bibliographystyle{icml2021}
\input{appendix}


%



\end{document}

%% file: defs.tex
\newcommand{\notes}[1]{}

\theoremstyle{definition}

\theoremstyle{plain}

\newcommand{\vecz}{\ensuremath{\mathbf{z}}}
\newcommand{\vech}{\ensuremath{\bm{h}}\xspace}

\newcommand{\ith}[1]{\ensuremath{i^{{th}}}}

\newcount\permx
\newcount\permy
\def\permdot#1#2{
\permx=#1 \advance\permx by-1
\permy=#2 \advance\permy by-1
\psframe[fillcolor=black, fillstyle=solid]
(\permx,\permy)(#1, #2)
}

\newcommand{\boxnum}[1]{{\setlength{\fboxsep}{1pt}\raisebox{1pt}{\hspace{1pt}\fbox{\tiny #1}\hspace{1pt}}}}
\newcommand{\ind}[1]{\ensuremath{_{\kern-0.5pt\boxnum{#1}}}}

\newcommand{\vecx}{\ensuremath{{\bm{x}}}\xspace}

\newcommand{\vecy}{\ensuremath{{\bm{y}}}\xspace}

\newcommand{\smallnt}[1]{\ensuremath{_{\mbox{\tiny PP}}}\xspace}

\newcommand{\pseudocode}{Algorithm}
\floatname{algorithm}{\pseudocode}

\iffalse

\else

\fi

\newcommand{\eos}{\mbox{\scriptsize \texttt{<eos>}}\xspace}

%% file: intro.tex

Speech-to-text translation (ST), which translates the source language speech
to target language text, is useful in many 
 scenarios such as international conferences, 
 travels, foreign-language video subtitling, etc.
Conventional cascaded approaches to ST \cite{Ney99,Matusov2005OnTI,Mathias2006,Berard2016} 
first transcribe the speech audio 
into source language text (ASR) 
and then perform text-to-text machine translation (MT),
which inevitably suffers from error propagation in the pipeline.
To alleviate this problem,
recent efforts explore end-to-end approaches (E2E-ST) 
\cite{Weiss2017, Berard2018EndtoEndAS, Vila2018EndtoEndST, Gangi2019},
which are computationally more efficient at inference time and mitigate the risk of 
error propagation from imperfect ASR.

\begin{figure}[!t]
\centering
\includegraphics[width=.8\linewidth]{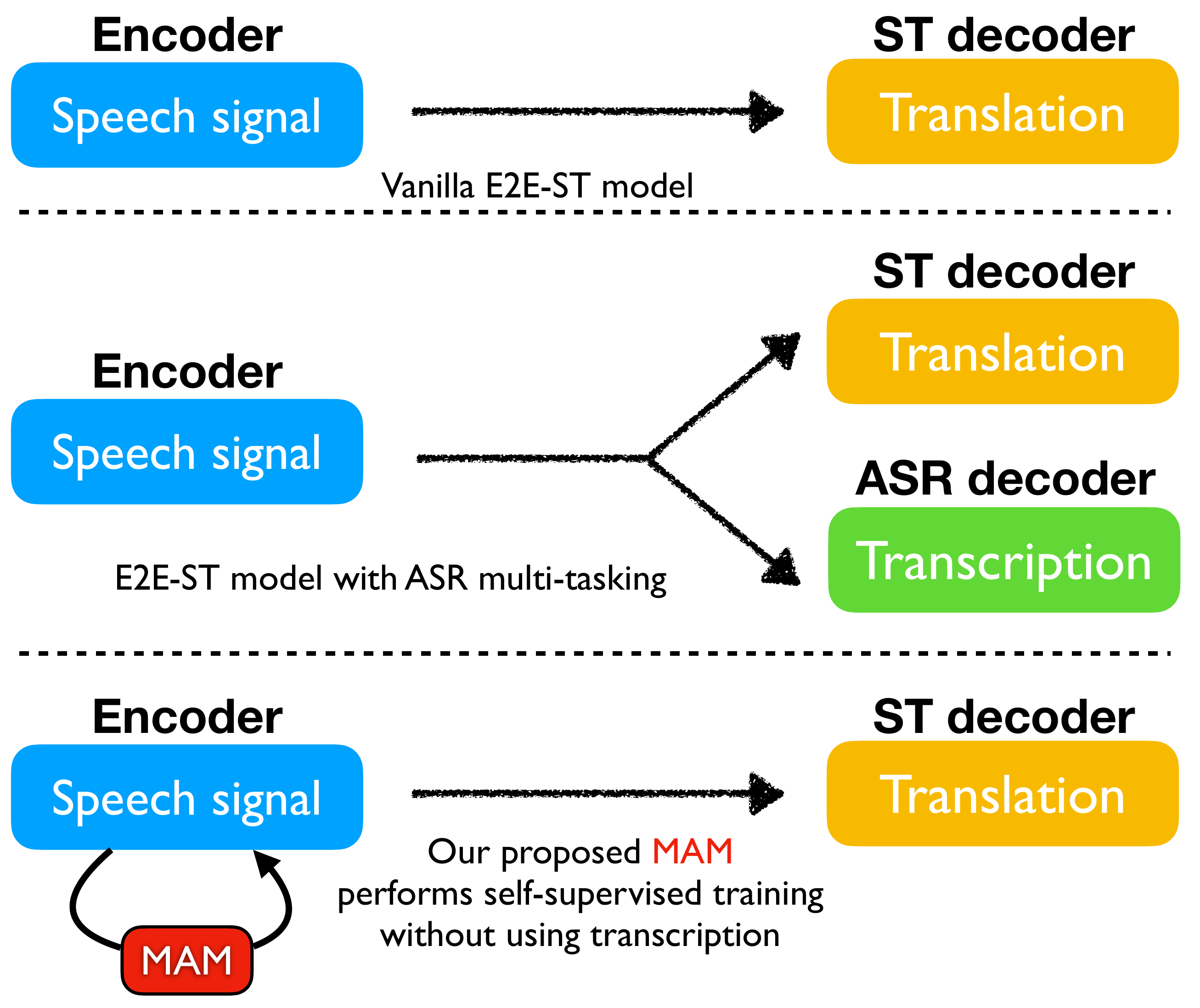}
\caption{Comparisons with different existing solutions and our proposed Masked Acoustic Modeling (MAM).}
\label{fig:comparison}
\vspace{-15pt}
\end{figure}

To improve the translation accuracy of E2E-ST models,
researchers either initialize the encoder of ST with 
a pre-trained 
ASR encoder  \cite{Berard2018EndtoEndAS,bansal2019,WangWLY020}
to get better representations of the speech signal,
or perform  Multi-Task Learning (MTL) with ASR to
bring more training and supervision signals to the shared encoder \cite{anastasopoulos2016,
anastasopoulos2018,
Sperber2019, Liu2019} (see Fig.~\ref{fig:comparison}).
These methods improve the translation quality by providing
more training signals to the encoder to learn better 
phonetic information and hidden 
representation correspondence \cite{Stoian2020}.

However, both above solutions assume the existence of 
substantial speech transcriptions of the source language.
Unfortunately,
this assumption
is problematic.
On the one hand, for certain low-resource languages, 
especially endangered ones \cite{bird2010,bird2014}, 
the source speech transcriptions 
are expensive to collect.
Moreover, according to the report from~\citet{Ethnologue},
there are more than 3000 languages that have no written form
or no standard orthography, 
making phonetic transcription impossible \cite{duong2016}.
On the other hand, 
the amount of speech audios with 
transcriptions are limited (as they are expensive to collect),
and there exist far more 
audios without any annotations.
It will be much more straightforward and cheaper to leverage
these raw audios to train a robust encoder directly.

To relieve from the dependency on source language transcriptions,
we present a straightforward yet effective solution, Masked Acoustic Modeling (MAM),
to utilize the speech data  in a self-supervised fashion
without using any source language transcription,
unlike other speech pre-training models \cite{Chuang2019,WangWLY020ACL}.
Aside from the regular training of E2E-ST 
(without ASR as MTL or pre-training),
MAM masks certain portions of the speech input randomly
and aims to recover the masked speech signals with their context
on the encoder side.
MAM not merely provides an alternative solution to improving E2E-ST,
but also is a general technique that 
can be used as a pre-training module on arbitrary acoustic signals,
e.g., multilingual speech, music, animal sounds.
The contributions of our paper are as follows:
\vspace{-5pt}
\begin{itemize}
\vspace{-5pt}
\item We demonstrate the importance of a self-supervising module for E2E-ST. Unlike all previous attempts, 
which heavily depend on transcription, 
MAM improves the capacity of the encoder by recovering masked speech signals merely based on their context. 
MAM also can be used together with transcriptions in 
ASR pre-training and MTL settings to further boost the translation accuracy.
\vspace{-5pt}
\item MAM also can be used as a pre-training module solely by itself. 
During pre-training, MAM is capable to utilize arbitrary acoustic signal
(e.g., music, animal sound) other than regular speech audio.
Considering there are much more acoustic data 
than human speech, MAM has better potential to be used for pre-training.
To the best of our knowledge, MAM is the first technique that is able to
perform pre-training with any form of the audio signal.
\vspace{-5pt}
\item For 8 different translation directions, 
when we do not use any transcription, MAM demonstrates an average 
BLEU improvements of 
1.09 in the basic setting and 2.26 with pre-training.
\vspace{-5pt}
\item We show that the success of MAM does not rely on intensive or 
expensive computation. MAM only has 6.5\% more parameters than the baseline model.
\vspace{-5pt}
\end{itemize}

%% file: prelim.tex
We first briefly review the standard E2E-ST and E2E-ST with ASR MTL
to set up the notations.

\vspace{-5pt}

\subsection{Vanilla E2E-ST Training with Seq2Seq}
Regardless of particular design of Seq2Seq models for different tasks,
the encoder always takes the source input sequence
$\vecx = (x_1,...,x_n)$ of $n$ elements 
where each $x_i \in \mathbb{R}^{d_x}$ is a $d_x$-dimension vector
and produces a sequence of hidden representations 
$\vech =f(\vecx) = (h_1,...,h_n)$
where $h_i = f(\vecx)$.
The encoding function $f$ can be implemented by 
a mixture between Convolution, RNN and Transformer.
More specifically, $\vecx$ can be the spectrogram or mel-spectrogram
of the source speech,
and each $x_i$ represents the frame-level
speech feature with certain duration.

On the other hand, the decoder greedily predicts a new output word
$y_t$ given both the source sequence \vecx
and the prefix of decoded tokens, denoted $\vecy_{<t}=(y_1,...,y_{t-1})$.
The decoder continues the generation until
it emits \eos and finishes the entire decoding process.
Finally, we obtain the hypothesis
$\vecy = (y_1,...,\eos)$ with the model score which
defined as following:
\begin{equation}
p(\vecy \mid \vecx) = \textstyle\prod_{t=1}^{|\vecy|}  p(y_t \mid \vecx,\, \vecy_{<t})
\label{eq:gensentscore}
\end{equation}

During the training time, the entire model
aims to maximize the conditional probability of each ground-truth target
sentence $\vecy^\star$ given input \vecx
over the entire training corpus $D_{\vecx, \vecy^\star}$, or equivalently minimizing the following loss:
\begin{equation}
\ell_{\text{ST}}(D_{\vecx, \vecy^\star}) = - \textstyle\sum_{(\vecx,\vecy^\star)\in D_{\vecx, \vecy^\star}} \log p(\vecy^\star \mid \vecx) 
\label{eq:train}
\end{equation}

\subsection{Multi-task Learning with ASR}

To further boost the performance of E2E-ST,
researchers proposed to either 
use pre-trained ASR encoder to initialize ST encoder,
or to perform ASR MTL together with ST training.
We only discuss the MTL since
pre-training does not require significant change to Seq2Seq model.

During multi-task training, there are two decoders 
sharing one encoder.
Besides the MT decoder, there is also another decoder for generating transcriptions.
With the help of ASR training,
the encoder is able to learn more accurate speech 
segmentations (similar to forced alignment)
making the global reordering of those segments for MT relatively easier.
We defined the following training loss for ASR:
\begin{equation}
\ell_{\text{ASR}}(D_{\vecx, \vecz^\star}) = - \textstyle\sum_{(\vecx,\vecz^\star)\in D_{\vecx, \vecz^\star}} \log p(\vecz^\star \mid \vecx) 
\label{eq:trainASR}
\end{equation}
where $\vecz^\star$ represents the annotated, ground-truth transcription 
for speech audio $\vecx$. 
In our baseline setting, we also hybrid CTC/Attention
framework \cite{Watanabe2017} on the encoder side.
In the case of multi-task training with ASR for ST, the total loss
is defined as 
\begin{equation}
\ell_{\text{MTL}}(D_{\vecx, \vecy^\star, \vecz^\star}) = \ell_{\text{ST}}(D_{\vecx, \vecy^\star})+\ell_{\text{ASR}}(D_{\vecx, \vecz^\star})
\label{eq:totalmlt}
\end{equation}
where $D_{\vecx, \vecy^\star, \vecz^\star}$ is the training dataset which 
contains speech, translation and transcription triplets.


\begin{figure*}[ht]
\centering
\includegraphics[width=.8\linewidth]{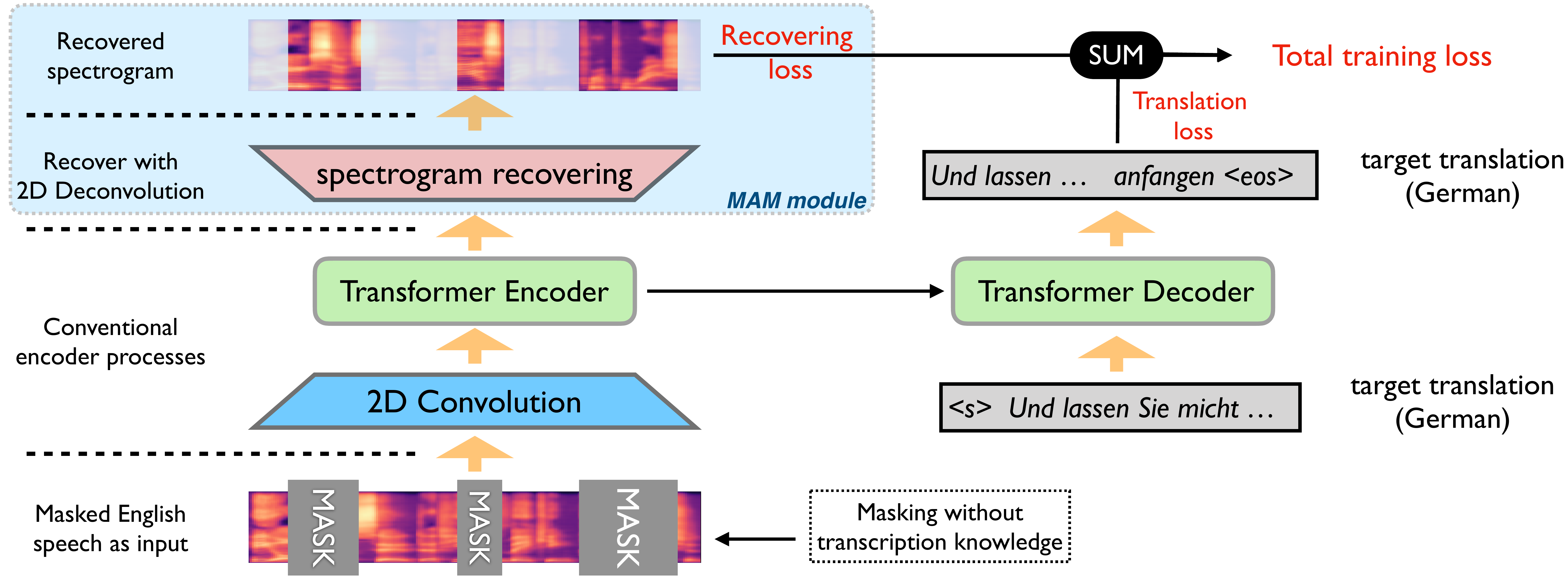}
\caption{MAM (in blue box) can be treated as one extra module besides 
standard Transformer encoder-decoder and
convolution layers for processing speech signals.}
\label{fig:mam}
\vspace{-5pt}
\end{figure*}
\vspace{-5pt}

%% file: methods.tex

All the existing solutions to boost the current E2E-ST performance
heavily depend on the availability of the transcription of the source language.
Those solutions are not able to take advantage
of large amount of speeches without any annotations.
They also become inapplicable when the source language
is low-resource
or even does not have a standard orthography system.
Therefore, the ideal solution should not be constrained by
source language transcription
and still achieves similar translation quality.
Thus, we introduce MAM in this section.

\subsection{MAM as Part of Training Objective}

We propose to perform self-supervised training
on the encoder side by reconstructing sabotaged speech signals from the input.
Note that MAM is totally different from another self-supervised training
\cite{Chuang2019,WangSemantic2020,WangWLY020ACL} which rely on transcription
to segment the speech audio with 
forced alignment tools\cite{Povey11thekaldi,McAuliffe2017}.
We directly apply random masks with different widths over speech audio,
eliminating the dependency of transcription.
Therefore, MAM can be easily applied to speech audio without transcription
and even to any non-human speech audio, e.g., music and animal sound.

Formally, we define a random replacement function over the 
original speech input $\vecx$:
\begin{equation}
\vspace{-5pt}
\hat{\vecx} \sim  \text{Mask}_{\text{frame}}(\vecx, \lambda),
\label{eq:randx}
\end{equation}
where $\text{Mask($\cdot$)}_{\text{frame}}$ randomly replaces some certain frames in $\vecx$
with the same random initialized vector, $\epsilon \in \mathbb{R}^{d_x}$,
 with a probability of $\lambda$ (30\% in our experiments).
Note that we use the same vector $\epsilon$ to represent
all the corrupted frames (see one example 
in Fig.\ref{fig:masked}).
Then we obtain a corrupted input $\hat{\vecx}$ and its corresponding 
latent representation $\hat{\vech}$.

For MAM module, we have the following training objective to reconstruct the 
original speech signal with the surrounding context 
information with self-supervised fashion:
\begin{equation}
\ell_{\text{Rec}}(D_{\vecx})= \textstyle\sum_{\vecx \in D_{\vecx}} || \vecx - \phi (f(\hat{\vecx})) ||_2^2
\label{eq:reconstruct}
\end{equation}
where $\phi$ is a reconstruction function which tries to recover the original
signal from the hidden representation $f(\hat{\vecx})$ with corrupted inputs.
For simplicity, we use regular 2D deconvolution as $\phi$, and
mean squared error
for measuring the difference between original input and recovered signal.
Finally, we have the following total loss of our model
\vspace{-5pt}
\begin{equation*}
    \vspace{-5pt}
\ell_{\text{MAM}}(D_{\vecx, \vecy^\star}) = \ell_{\text{ST}}(D_{\vecx, \vecy^\star})+\ell_{\text{Rec}}(D_{\vecx})
\label{eq:totalloss}
\end{equation*}

To further boost the performance of E2E-ST, 
we can train MAM with ASR MTL when 
transcription is available:
\vspace{-5pt}
\begin{equation*}
    \vspace{-5pt}
\ell_{\text{MAM + MTL}}(D_{\vecx, \vecy^\star, \vecz^\star}) = \ell_{\text{MTL}}(D_{\vecx, \vecy^\star, \vecz^\star})+\ell_{\text{Rec}}(D_{\vecx})
\label{eq:totallossmlt}
\end{equation*}

\subsection{Different Masking Strategies}
\label{sec:seg}


MAM aims at much harder tasks than pure textual pre-training models,
e.g., BERT or ERINE, which only perform semantic learning over missing tokens.
In our case, we not only try to recover semantic meaning, but also acoustic 
characteristics of given audio.
MAM simultaneously predicts the missing words and generates spectrograms like 
speech synthesis tasks.  

To ensure the masked segments contain different levels of granularity of 
speech semantic,
we propose the following masking methods.
\vspace{-5pt}
\paragraph{Single Frame Masking} Uniformly mask $\lambda \%$ frames out of $\vecx$ to 
construct $\hat{\vecx}$. Note that we might have continuous frames that were masked.
\vspace{-5pt}
\paragraph{Span Masking} Similar with SpanBERT~\cite{joshi2020spanbert}, we first sample a serial of span widths and then apply those spans randomly to different positions of the input signal. Note that we do not allow 
overlap in this case.
Our span masking is defined as $\hat{\vecx} \sim  \text{Mask}_{\text{span}}(\vecx, \lambda)$.





\subsection{Pre-training MAM}

MAM is a powerful technique that is not only beneficial 
to the conventional training procedure,
but also can be used as a pre-training framework that does not need any annotation.

The bottleneck of current speech-related tasks, e.g., ASR, ST, is lacking of 
the annotated training corpus. 
For some languages that do not even have a standard orthography system, 
these annotations are even impossible to obtain.
Although current speech-related, pre-training frameworks \cite{Chuang2019,WangWLY020ACL}
indeed relieve certain needs of large scale parallel training corpus for E2E-ST,
all of these pre-training methods still need intense transcription 
annotation for the source speech.

During pre-training time, we only use the encoder part of MAM.
Thanks to our flexible masking techniques, 
MAM is able to perform pre-training with any kind of audio signal.
This allows us to perform pre-training with MAM with 
three different settings, pre-training with source language speech,
with multilingual speech, and arbitrary audios.
To the best of our knowledge, 
MAM is the first pre-training technique that can be applied 
to arbitrary audios. 
Considering about the vast arbitrary acoustic signals existing
on the Internet (e.g., youtube),
MAM has great potential to further boost 
the downstream tasks.
MAM that pre-trained with arbitrary acoustic signals
does not differentiate languages and provides the unified pre-trained 
model for any downstream, fine-tuning task.
This is different from the multilingual pre-training setting since
the downstream task's source speech language is not necessary
to be included in the pre-training stage,
which is essential to the low-resource and zero-resource languages.


%% file: exp.tex
In this section, 
we conducted MAM pre-training experiment on three corpora, 
Libri-Light (only English speech, medium version)~\cite{librilight},
Common Voice~\footnote{https://commonvoice.mozilla.org/en/datasets} (We select 14 languages, which contains 
ca, de, en, es, fr, it, kab, nl, pl, pt, ro, ru, zh-CN, zh-TW)~\cite{ardila2020common},
and Audioset (arbitrary acoustic data)~\cite{audioset}.
The statistical results of the dataset are shown in Table.~\ref{tab:corpora}.
Note that Audioset includes a wide range of arbitrary sounds, from human and animal sounds, to natural and environmental sounds, to musical and miscellaneous sounds.

Then, we analyze the performance of MAM 
in E2E-ST with 8 different language 
translation directions using 
English as the source speech on MuST-C dataset~\cite{mustc}.
All raw audio files are processed by Kaldi~\cite{Povey11thekaldi}
to extract 80-dimensional log-Mel filterbanks stacked with 3-dimensional pitch features
using a window size of 25 ms  and step size of 10 ms.
We train sentencepiece~\cite{kudo-richardson-2018-sentencepiece} 
models with a joint vocabulary size of 8K for each dataset. 
We remove samples that have more than 3000 frames for GPU efficiency.
Our basic Transformer based E2E-ST framework has similar settings 
with ESPnet-ST\cite{inaguma2020espnet}. 
We first downsample the speech input with 2 layers of 
2D convolution of size 3 with stride size of 2.
Then there is a standard 12-layers Transformer
with 2048 hidden size to
bridge the source and target side.
We only use 4 attention heads on each side of the transformer 
and each of them has a dimensionality of 256.
For MAM module, we simply linearly project the outputs of the Transformer
encoder to another latent space, then upsample the latent representation
with 2-layers deconvolution to match the size of the original input signal.
For the random masking ratio $\lambda$, we choose 30\% across all the 
experiments including pre-training. 
During inference, we do not perform any masking over the speech input.
We average the last 5 checkpoints for testing.
For decoding,
we use a beam search with setting beam size and length penalty to 5 and 0.6, respectively.

\begin{table}[]
\centering
\resizebox{0.8\linewidth}{!}{
\begin{tabular}{cccc}
\toprule
                                                          & ST  & ST+ASR & ST+MAM \\ \midrule
\begin{tabular}[c]{@{}c@{}}\# of parameters\end{tabular} & 31M & 47M    & 33M    \\ \bottomrule
\end{tabular}
}
\caption{MAM only has 6.5\% more parameters than the baseline model while ASR 
multi-tasking needs to use 51.6\% more parameters.}
\label{tab:parameters}
\end{table}

\begin{table}[]
  
  \setlength{\tabcolsep}{0.3em}
   \resizebox{\linewidth}{!}{
   \begin{tabular}{|c|c|c|c|c|c|}
   \hline
         & MuST-C  & Libri-Speech & Libri-Light & Common Voice & Audioset \\ \hline
   Type  & $\blacklozenge$$\bigstar$ & $\blacklozenge$ &  &$\blacklozenge$ & \\\hline
   Hours & 408h & 960h    & 3748h  & 4421h & 6873h \\ \hline
   \end{tabular}
   }
\caption{The statistical results of corpora. 
$\blacklozenge$ and $\bigstar$  denote the corpus has transcripts and translations, respectively.
Note that although Common Voice has transcripts, we do not use them.}
\label{tab:corpora}
\end{table}

Our MAM is very easy to replicate as we do not perform any parameters 
and architecture search upon the baseline system. 
Due to the simple, but effective design of MAM,
MAM does not rely on intensive computation.
It converges within 2 days of training with 8 1080Ti GPUs for 
the basic model.
We showcase the comparison of parameters between different solutions 
to E2E-ST in Table.~\ref{tab:parameters}
This makes a big difference with current popular intensive computations frameworks
such as BERT\cite{BERT} (340M parameters) and GPT3\cite{brown2020language} (175B parameters),
making this technique is accessible to regular users.

\subsection{Analyzing ASR and MAM}
\label{sec:anaylsis}

\begin{figure}[!t]
\centering
\includegraphics[width=.8\linewidth]{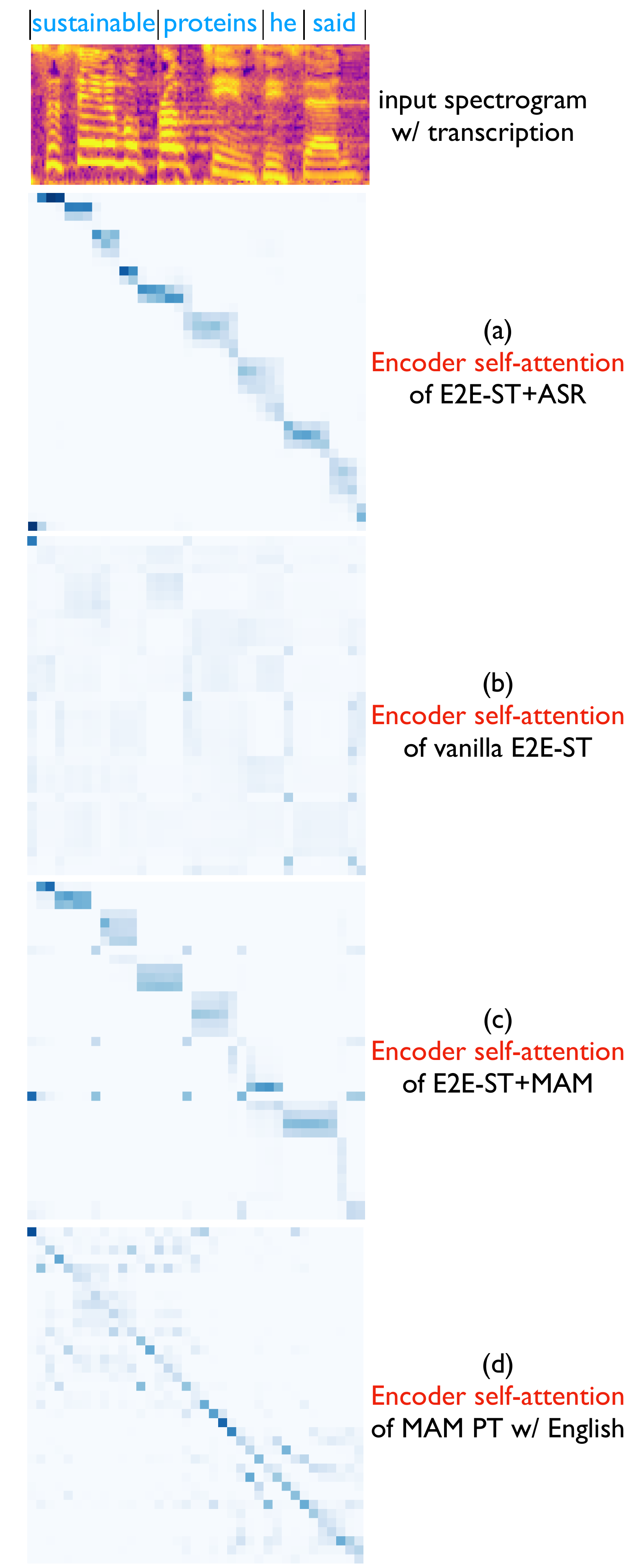}
\caption{One head of the last layer self-attention
comparison between different models.
 ASR MTL and MAM help the encoder learns similar self-attentions. See detailed discussion in \ref{sec:anaylsis}.}
\label{fig:STvsASR}
\end{figure}

Aside from the extra training signal that is introduced
by transcriptions, there is a deeper reason why 
ASR and MAM are beneficial to E2E-ST.
In this section, we first discuss the difficulties 
and challenges in E2E-ST.
Then we analyze the reasons why ASR MTL and MAM 
are helpful for E2E-ST by visualizing the self-attention over
source side encoder.

Compared with other tasks, e.g., MT or ASR, which also
employ Seq2Seq framework for E2E training,
E2E-ST is a more difficult and challenging task in many ways.
Firstly, data modalities are different 
on the source and target sides. 
For ST, the encoder deals with speech signals and tries to learn
word presentations on the decoder side, while MT has 
text format on both sides.
Secondly, due to the nature of the high sampling rate of speech signals,
speech inputs are generally multiple (e.g. 4 to 7)  
times longer than the target sequence,
which increases the difficulties of learning 
the correspondence between source and target.
Thirdly, compared with the monotonicity natural of the
alignment of ASR, ST usually needs to learn the global reordering
between speech signal and translation, and this raises the difficulties
to another level.
Especially in ST, since source and target are in different languages,
it is very challenging to obtain the corresponding phoneme or syllable
segments given the training signal from a different language.

Fig.~\ref{fig:STvsASR} tries to explain and analyze
the difference between
E2E-ST (a) and E2E-ST with ASR MTL (b).
We extract the most top layer from the encoder for comparison.
We notice that E2E-ST (a) tends to get more meaningful self-attention on the encoder
with the training signal from ASR. 
With help from ASR, 
the source input spectrogram is chunked into segments
that contain phoneme-level information.
During training, the monotonicity natural of the
ASR alignment functions as a forced alignment to group a set of adjacent frames
to represent certain phonemes or syllables from source speech.
With a larger scale of segmented spectrograms,
the target side decoder only needs to perform reordering on 
those segments instead of frames.
Our observations also align with 
the analysis from \citet{Stoian2020}.

We also visualize the self-attention on encoder for E2E-ST with MAM 
(without pre-training) in (c) of Fig.~\ref{fig:STvsASR}. 
We find that MAM has the similar ability with ASR to segment
the source speech into chunks.
As it is shown in (d) of Fig.~\ref{fig:STvsASR},
when we only perform pre-training on the 
English speech (Libri-Light dataset), without
E2E-ST training, 
self-attentions that are generated by pre-trained MAM 
are mostly monotonic on source side.
Recovering local frames usually needs 
the information from surrounding context, especially for the speaker and environment-related characteristic.
But we still observe that self-attention sometimes focuses
on longer distance frames as well.
This type of attention is very similar with low to mid layer self-attention
of ASR.
When there is a down streaming task (e.g., ASR or ST) is used for fine tuning,
the top layer's self-attention will get chunked attention which is similar to (a) and (c).

To conclude, we observe that MAM functions very similar to ASR
on the encoder side.
Hence, MAM is a reliable framework that can be used as 
an alternative solution when there is no transcription available.
Especially, with the help of a large scale acoustic dataset, which
does not have transcription annotation, 
MAM provides the E2E-ST a much better encoder initialization.

\if
However, the above benefits of ASR are introduced by utilizing 
extra tremendous effort to annotate the corresponding transcription.
And sometimes the transcription is hard or impossible to collect for 
the languages without written form or no standard orthography.
To relive from the dependence of using transcription, we instead to propose MAM which 
is capable of 
achieving similar functionality of ASR pre-train and MTL in E2E-ST 
without using transcriptions 
via a self-supervised training fashion in 
both conventional E2E-ST training 
and pre-training stages.
When the encoder of MAM is trained to predict the missing speech chunks,
MAM firstly learns to interpret the observable speech part
and then fills the blanks frames with both semantic and speech information.
Through this procedure, MAM automatically learns the segmented chunks 
of self-attention over source side.
\fi

\subsection{Visualizing Reconstruction}

\begin{figure}[h]
\centering
\begin{subfigure}[b]{0.44\textwidth}
   \includegraphics[width=1\linewidth]{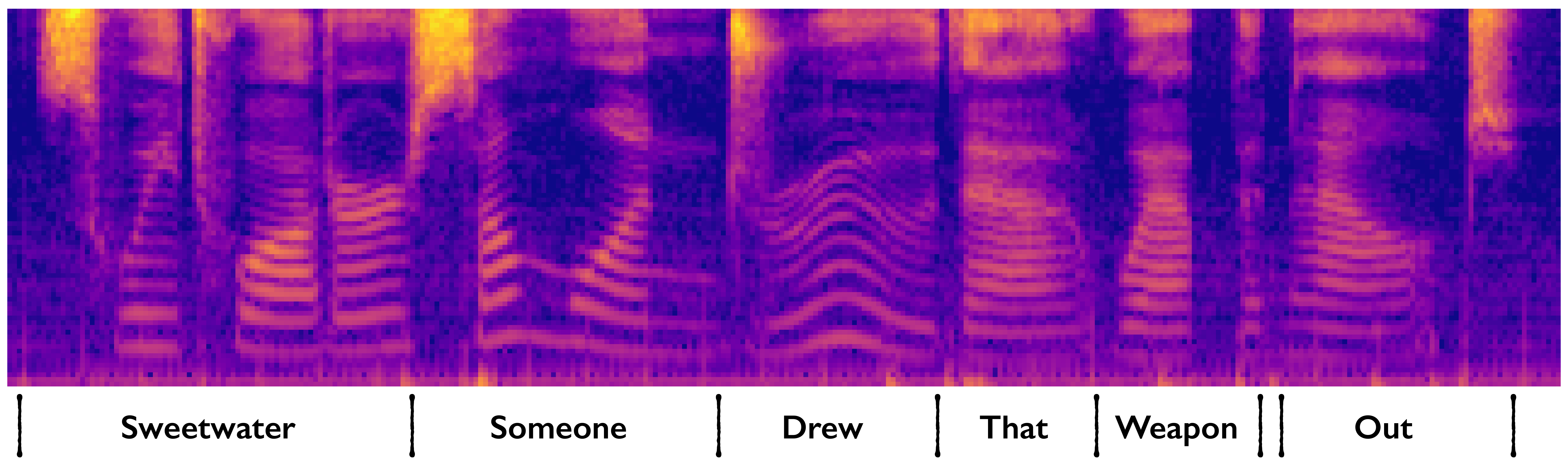}
   \caption{The original speech spectrogram. Note that though we annotate the transcription underneath, we do not use transcription information at all during pre-training.}
   \label{fig:original}
\end{subfigure}
\begin{subfigure}[b]{0.44\textwidth}
   \includegraphics[width=1\linewidth]{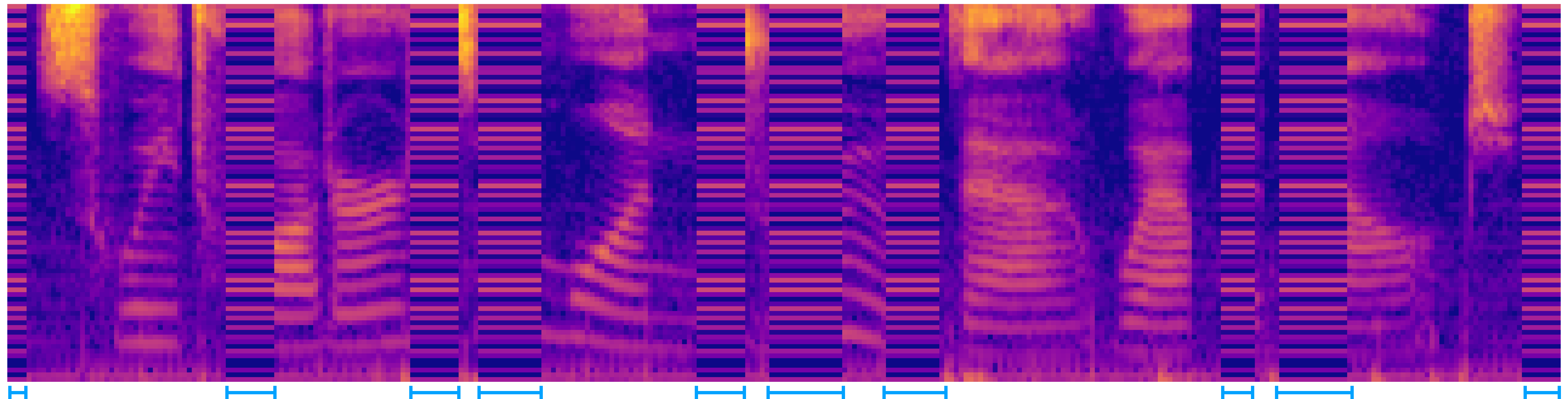}
   \caption{We mask the selected frames (underlined with blue lines) with the same random initialized vector.}
   \label{fig:masked}
\end{subfigure}
\begin{subfigure}[b]{0.44\textwidth}
   \includegraphics[width=1\linewidth]{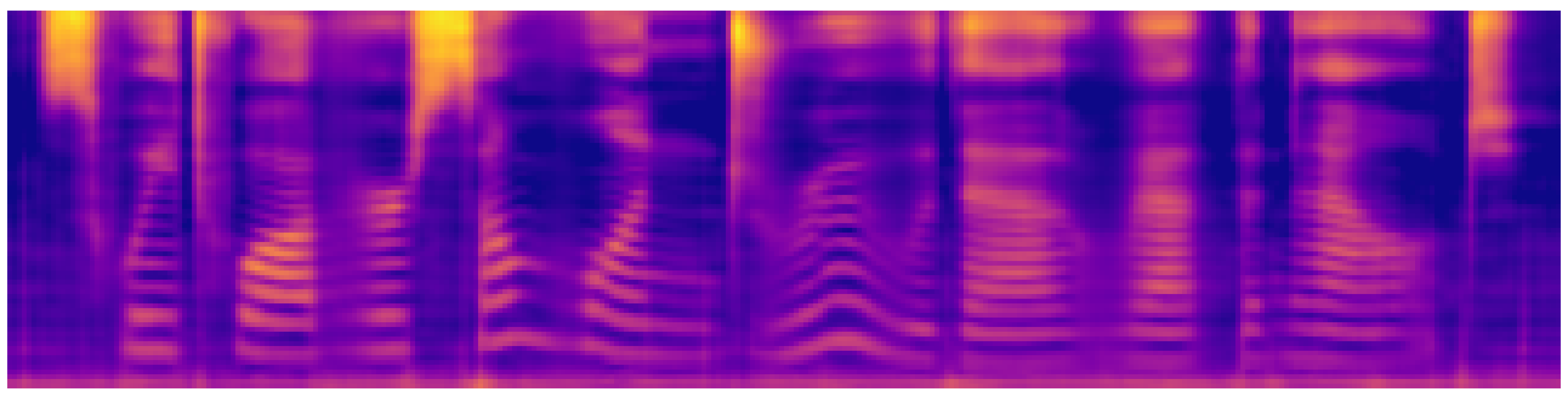}
   \caption{Recovered spectrogram with MAM, pre-trained with Libri-Light corpus.}
   \label{fig:libri_rec}
\end{subfigure}
\begin{subfigure}[b]{0.44\textwidth}
   \includegraphics[width=1\linewidth]{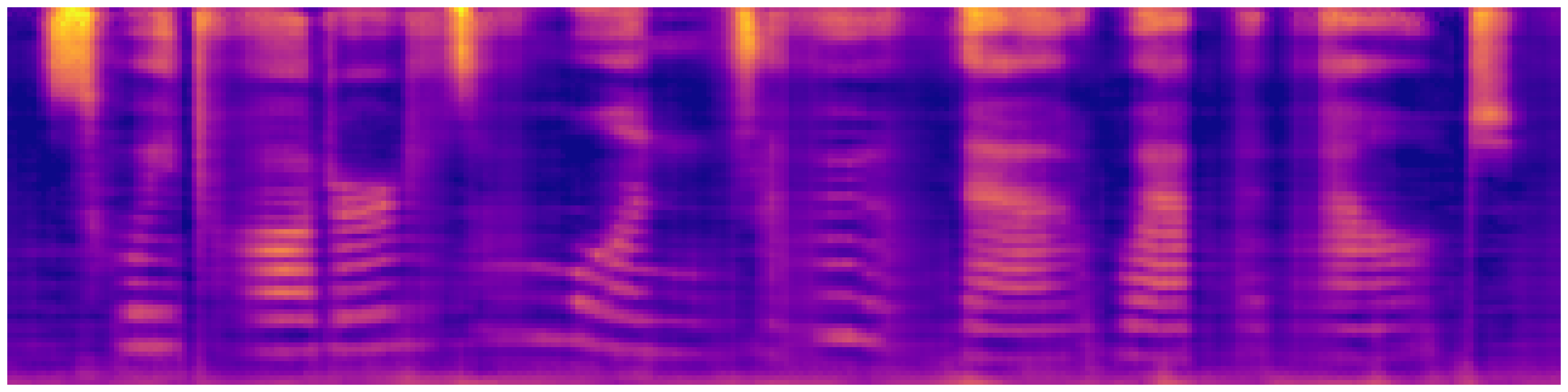}
   \caption{MAM that pre-trains with FMA music corpus still have the ability to reconstruct corrupted speech signal.}
   \label{fig:fma_rec} 
\end{subfigure}
\caption{One speech example to showcase the reconstruction ability of pre-trained MAM. We notice that MAM reconstructs the corrupted audio signal in both pre-training with ordinary speech and music dataset.}
\label{fig:rec}
\end{figure}

To demonstrate what MAM has learned from pre-training step, 
we first showcase the reconstruction
ability of MAM by visualizing the differences of spectrograms
between the original and recovered inputs.
This experiment was conducted on two corpora,
Libri-Light and the Free Music Archive (FMA)~\cite{defferrard2016fma} dataset.
We use the ``fma-medium'' setting \footnote{https://github.com/mdeff/fma}
which contains about 25,000 tracks of 30 seconds 
music within 16 unbalanced genres.
The total music length is about 208 hours.
We use FMA dataset for reconstruction visualization since
FMA only contains music data and the characteristic of the music signal
is very different from pure human speech.
Note that our reconstructed spectrograms are a little blur compared with
the original input since there are some downsampling steps in the E2E-ST baseline 
framework.

To verify the pre-trained results of MAM, we demonstrate the reconstruction
ability of MAM by visualizing the results in Fig.~\ref{fig:rec}.
We show the original spectrogram of input speech in 
Fig.~\ref{fig:original}.
Then we corrupted the original spectrogram by replacing the selected 
mask frames with $\epsilon$, which is a random initialized vector,
to form $\hat{\vecx}$ (see Fig.~\ref{fig:masked}).
In Fig.~\ref{fig:libri_rec}, we show that our proposed MAM is able to recover
the missing segments of input speech
by pre-training over the Libri-Light dataset.
More interestingly, 
since MAM does not need any transcription to perform pre-training,
we also pre-train MAM with FMA dataset.
Surprisingly, as shown in Fig.~\ref{fig:fma_rec},
MAM performs very similar reconstruction ability 
compared with the one that is pre-trained with speech dataset
considering the corrupted audio is only about pure speech.
This might be because some music tracks include
human singing voices and MAM learns human speech 
characteristics from those samples though
human singing voice can be quite different from speech.
We also conduct reconstruction with speech pre-trained
MAM for corrupted FMA data (see Fig.~\ref{fig:recfma} in Appendix).

\if
In the other way around, 
we also try to use Libri-Light pretrained MAM
to recover the corrupted music in Fig.~\ref{fig:recfma}.
MAM that pre-trained with human speech data
does not show good reconstruction in Fig.~\ref{fig:fma2_rec}
since there are many different musical 
instruments' sounds that are unseen in speech data.

\begin{figure}[h]
\centering
\begin{subfigure}[b]{0.48\textwidth}
   \includegraphics[width=1\linewidth]{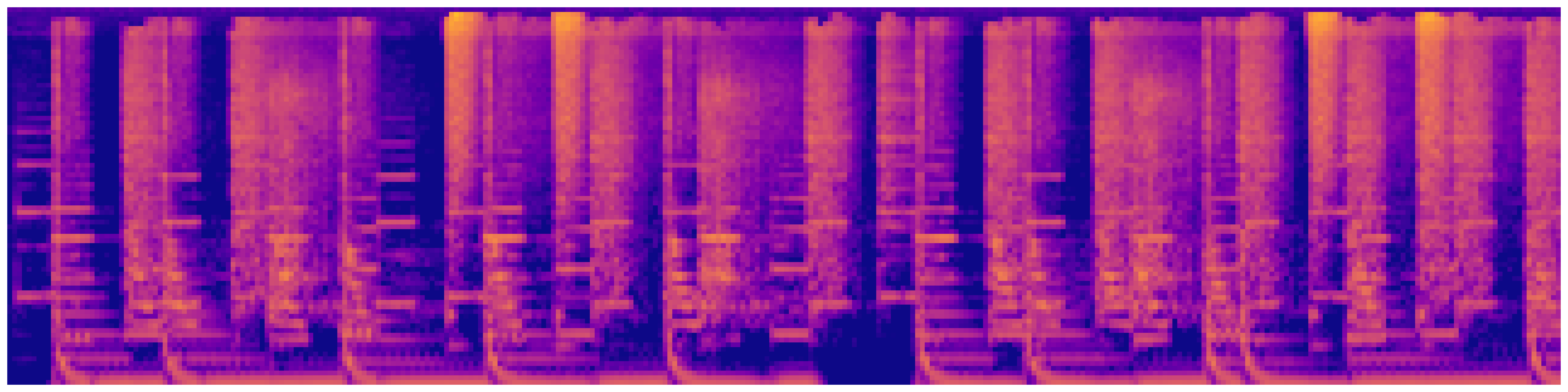}
   \caption{The original musical spectrogram that is mixed with different
   instruments' sound.}
   \label{fig:fmaoriginal}
\end{subfigure}
\begin{subfigure}[b]{0.48\textwidth}
   \includegraphics[width=1\linewidth]{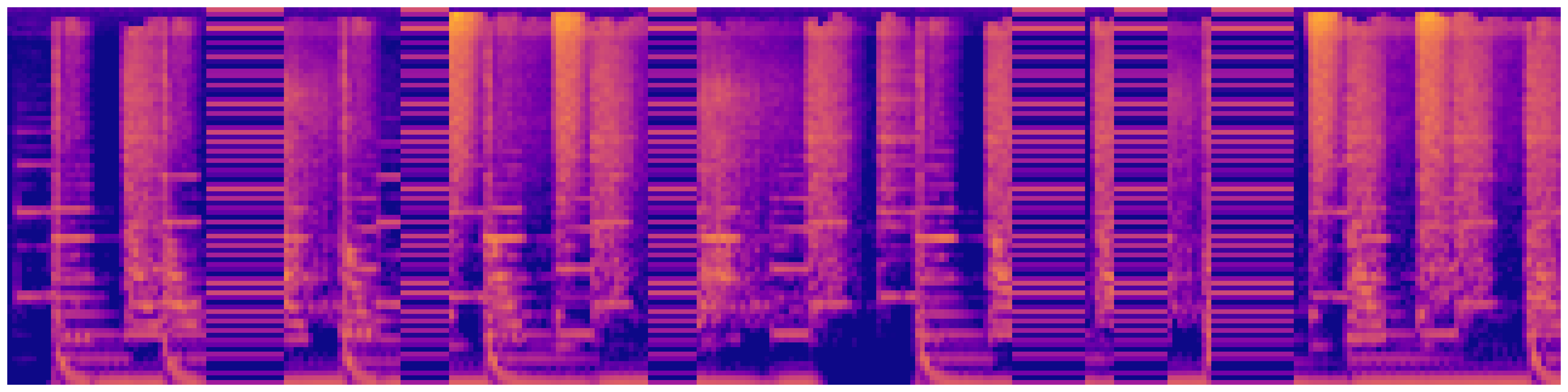}
   \caption{We mask the selected frames (underlined with blue lines) with the same random initialized vector.}
   \label{fig:fmamasked}
\end{subfigure}
\begin{subfigure}[b]{0.48\textwidth}
   \includegraphics[width=1\linewidth]{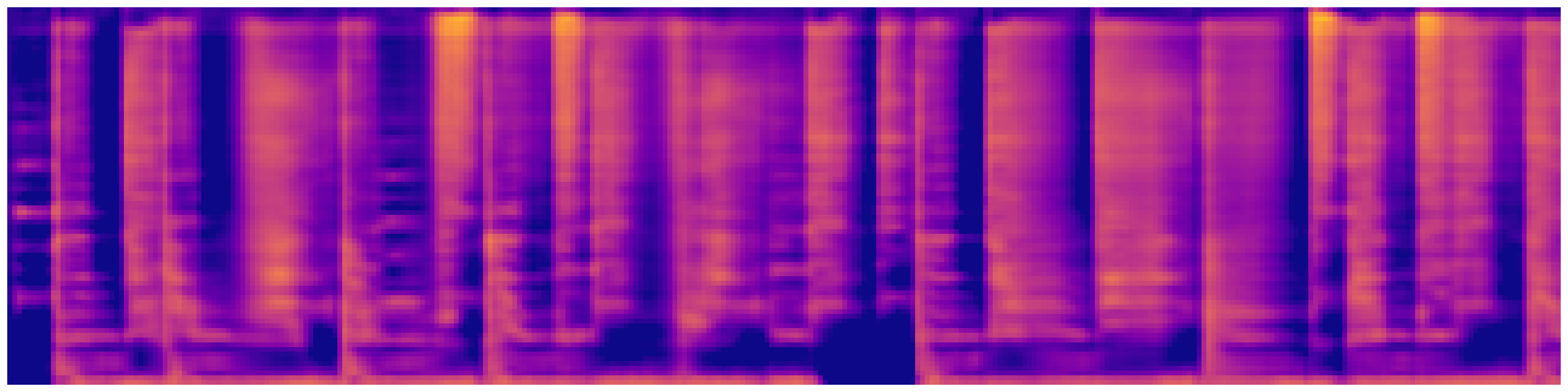}
   \caption{Recovered spectrogram with MAM, pre-trained with Libri-Light corpus.}
   \label{fig:fma2_rec}
\end{subfigure}
\caption{One speech example to showcase the reconstruction ability of pre-trained MAM. Pre-trained MAM with Libri-Light corpus (only human speech data)
can not reconstruct the original music spectrogram accurately since there are
many different musical instruments' sound that is unseen in speech data.}
\label{fig:recfma}
\end{figure}
\fi

\begin{table*}[th]
\centering
\resizebox{2\columnwidth}{!}{
\begin{tabular}{llrrrrrrrrr}
\hline
 &
  Models &
  \multicolumn{1}{c}{De} &
  \multicolumn{1}{c}{Es} &
  \multicolumn{1}{c}{Fr} &
  \multicolumn{1}{c}{It} &
  \multicolumn{1}{c}{Nl} &
  \multicolumn{1}{c}{Pt} &
  \multicolumn{1}{c}{Ro} &
  \multicolumn{1}{c}{Ru} &
  \multicolumn{1}{c}{Avg. $\Delta$} \\ \hline
\multirow{4}{*}{\rotatebox[origin=c]{90}{Baselines}}   
                             & MT with ASR annotation \cite{mustc} & 28.09 & 34.16 & 42.23 & 30.40 & 33.43 & 32.44 & 28.16  & 18.30 &  \\
                             & Cascaded methods \cite{inaguma2020espnet}        & 23.65 & 28.68 & 33.84 & 24.04 & 27.91 & 29.04 & 22.68  & 16.39 &  \\ 
                             & E2E-ST                 & 19.64 & 23.68 & 28.91 & 19.95 & 23.01 & 24.00 & 21.06  & 12.05 & \multicolumn{1}{c}{-} \\
                             & ST + SpecAug           & 20.06 & 24.51 & 29.26 & 20.27 & 23.73 & 24.40 & 21.21 & 12.84 & +0.49 \\ \hline \hline
\multirow{2}{*}{\rotatebox[origin=c]{90}{Ours}} & MAM (single)           & 20.34 & 24.46 & 29.18 & 19.52 & 23.81 & 24.56 & 21.37  & 12.57 & \textcolor{tgreen}{+0.44} \\
                             & MAM (span)             & 20.78 & 25.34 & 30.26 & 20.51 & 24.46 & 24.90 & 21.62  & 13.14 & \textcolor{tgreen}{\textbf{+1.09}} \\ \hline
\end{tabular}}
\caption{Comparisons between MAM and other baselines over 8 languages on MuST-C. 
In this setting, we use MAM as an extra training module for E2E-ST, and there is no pre-training involved.
We notice that MAM with span masking achieves better performance and there is 1.09 BLEU score improvements upon E2E-ST.
The column starts with ``Avg. $\Delta$'' summarizes 
the average improvements upon baseline method, E2E-ST.
See more discussions in Sec.~\ref{sec:withoutpt}.
}
\label{tb:withoutpt}
\end{table*}

\begin{table*}[th]
\centering
\resizebox{1.8\columnwidth}{!}{
\renewcommand{\arraystretch}{1.15}
\begin{tabular}{llrrrrrrrrr}
\hline
 &
  Models &
  \multicolumn{1}{c}{De} &
  \multicolumn{1}{c}{Es} &
  \multicolumn{1}{c}{Fr} &
  \multicolumn{1}{c}{It} &
  \multicolumn{1}{c}{Nl} &
  \multicolumn{1}{c}{Pt} &
  \multicolumn{1}{c}{Ro} &
  \multicolumn{1}{c}{Ru} &
  \multicolumn{1}{c}{Avg. $\Delta$} \\ \hline
\multirow{3}{*}{\rotatebox[origin=c]{90}{Baselines}}   & E2E-ST                 & 19.64 & 23.68 & 28.91 & 19.95 & 23.01 & 24.00 & 21.06 & 12.05 & \multicolumn{1}{c}{-} \\
                             & E2E-ST+ASR PT$^*$          & 20.75 & 25.57 & 30.75 & 20.62 & 24.31 & 25.33 & 22.50 & 14.24 & $^\dagger$+1.47 \\ 
                             & E2E-ST+ASR MTL         & 21.70 & 26.83 & 31.36 & 21.45 & 25.44 & 26.52 & 23.71 & 14.54 & $^\dagger$+2.41 \\ \hline \hline
\multirow{3}{*}{\rotatebox[origin=c]{90}{Ours}} & MAM w/ English PT      & 21.44 & 26.48 & 31.21 & 21.28 & 25.22 & 26.41 & 23.83 & 14.53 & \textcolor{tgreen}{\textbf{+2.26}} \\
                             & MAM w/ multilingual PT & 21.02 & 25.93 & 30.62 & 21.05 & 24.87 & 25.64 & 22.94 & 13.90 & \textcolor{tgreen}{+1.71} \\
                             & MAM w/ acoustic PT     & 20.81 & 25.85 & 30.48 & 20.52 & 24.81 & 25.46 & 22.90 & 13.83 & \textcolor{tgreen}{+1.55} \\ \hline
\end{tabular}}
\caption{Comparisons between span-based MAM pre-training with different pre-training corpus and other baselines over 8 languages on MuST-C. 
PT is short for pre-training. 
$^*$ denotes pretrained with Librispeech corpus.
We use Libri-Light for 
English pre-training, Common Voice for multi-lingual pre-training and 
Audioset for arbitrary acoustic data pre-training.
The methods denote with $\dagger$ use transcription in pre-training or MTL, but all our MAM methods do not use transcription.
MAM pre-training with English corpus achieves very similar performance with E2E-ST+ASR MTL.
The column starts with ``Avg. $\Delta$'' summarizes 
the average improvements upon baseline method.
See more discussions in Sec.~\ref{sec:withpt}.}
\label{tb:withpt}
\end{table*}

\begin{table*}[th]
\centering
\resizebox{1.7\columnwidth}{!}{
\begin{tabular}{lrrrrrrrrr}
\hline
Models &
  \multicolumn{1}{c}{De} &
  \multicolumn{1}{c}{Es} &
  \multicolumn{1}{c}{Fr} &
  \multicolumn{1}{c}{It} &
  \multicolumn{1}{c}{Nl} &
  \multicolumn{1}{c}{Pt} &
  \multicolumn{1}{c}{Ro} &
  \multicolumn{1}{c}{Ru} &
  \multicolumn{1}{c}{Avg. $\Delta$} \\ \hline
E2E-ST+ASR MTL                                                         & 21.70 & 26.83 & 31.36 & 21.45 & 25.44 & 26.52 & 23.71 & 14.54 & \multicolumn{1}{c}{-} \\ \hline \hline
MAM+ASR MTL                                                            & 22.41 & 26.89 & 32.55 & 22.12 & 26.49 & 27.22 & 24.45 & 14.90 & \textcolor{tgreen}{+0.69} \\
\begin{tabular}[c]{@{}l@{}}MAM w/ English PT \\ + ASR MTL\end{tabular} & 22.87 & 26.86 & 32.80 & 22.12 & 26.81 & 27.43 & 24.65 & 15.21 & \textcolor{tgreen}{\textbf{+0.90}} \\ \hline
\end{tabular}}
\caption{Comparisons between MAM with ASR MTL and E2E-ST with ASR MTL. MAM still achieves an improvement about +0.9 BLEU. The column starts with ``Avg. $\Delta$'' summarizes 
the average improvements upon baseline method.}
\label{tb:withmtl}
\end{table*}

\subsection{Translation Accuracy Comparisons}
\label{sec:compair}
We showcase the translation accuracy of MAM comparing against to
6 baselines from Table~\ref{tb:withoutpt} to Table~\ref{tb:withmtl}:
\vspace{-5pt}
\begin{itemize}
\item \textbf{Cascade}: cascade framework first transcribes the speech into
transcription then passes the results to later machines translation system.
\vspace{-5pt}
\item \textbf{MT with ASR annotation}: an MT system which directly generates the target 
translation from the human-annotated transcription.
\vspace{-5pt}
\item \textbf{E2E-ST}: this is the vanilla translation system which does not use transcriptions in MuST-C.
\vspace{-5pt}
\item \textbf{E2E-ST + ASR MTL}: ST trained with ASR MTL using the transcription in MuST-C.
\vspace{-5pt}
\item \textbf{ST + SpecAugment}: a data augmentation method ~\cite{Park_2019,Bahar:2019} by performing random masking over input speech.
\vspace{-5pt}
\item \textbf{E2E-ST + ASR PT}: the encoder of ST is initialized by a 
pre-trained ASR encoder
which is trained from the speech and 
transcription pairs in Libri-Speech~\cite{panayotov2015librispeech}.
\end{itemize}

To better make a conclusion of our results from Table~\ref{tb:withoutpt} to Table~\ref{tb:withmtl},
we organize the comparisons as follows.

\subsubsection{Comparison in the settings without transcription and pre-training}
\label{sec:withoutpt}
In Table~\ref{tb:withoutpt}, we first compare MAM against E2E-ST where there is no transcription and pre-training. 
Both MAM with single and span masking methods 
achieve averagely 
+0.44 (single) and +1.09 (span) 
improvements in BLEU score correspondingly against to E2E-ST
in 8 different translation directions.
Span masking consistently outperforms single frame masking 
as it is a more difficult self-supervised task.

\subsubsection{Comparison in the pre-training settings without transcription}
\label{sec:withpt}
In Table~\ref{tb:withpt}, we have three different pre-training settings for MAM,
 which are pre-training with 
English speech (same with the source language) from Libri-Light, 
multilingual speech data from Common Voice,
and arbitrary acoustic data from Audioset corpus.
Among those methods, MAM pre-trained with Libri-Light achieves the best
results as it consistently outperforms the baseline.
Averagely speaking, 
there is +2.26 improvements compared with E2E-ST.
When we compare to ``E2E-ST+ASR PT'',
there are about +0.79 improvements in BLEU score
across 8 target languages.

Surprisingly, 
MAM trained with acoustic data still achieves about +1.55 improvements upon E2E-ST.
Considering acoustic data does not need any annotation 
and this kind of dataset is much easier to collect,
the results are very encouraging.
With the help of vast acoustic data on the website (e.g., youtube),
MAM trained with arbitrary acoustic data has great potential to
further boost the performance.
To the best of our knowledge, MAM is the first technique that 
performs pre-training with any form of the audio signal.

MAM trained with Common Voice does not have significant improvements with 
two following reasons: firstly, speech audios in Common Voice
sometimes are very short (about 2 to 3 seconds) while
MuST-C usually contains much longer speech (above 10 seconds)
leading to very limited options for random masking; 
secondly there are much fewer English speech in this corpus.


\subsubsection{Comparison in the settings using transcription}
\label{sec:withmtl}
In this setting, we use ``E2E-ST + ASR MTL'' as the baseline.
MAM MTL with pre-training over Libri-Light achieves +0.9 average
improvements over 8 languages.

\subsubsection{Comparison to Wav2vec}
\label{sec:wav2vec}

We also compare MAM against to other wav2vec-based methods~\cite{wu+:2020} in Table~\ref{tb:wav2vec}.
Due to the differences in baseline methods, to make a fair comparison,
we only compare the relative improvements upon our own baseline on the same test data.
MAM still achieves much larger improvements upon a much stronger baseline.
Especially our baseline is already about 4 BLEU points better than the baseline in wav2vec,
MAM still achieves +1.6 more BLEU points improvements compared with wav2vec-based pre-training methods
making our performance on En-Ro about 5.6 BLEU points better than wav2vec-based methods.

\begin{table}[]
\centering
\resizebox{0.9\columnwidth}{!}{
\begin{tabular}{lrrrr}
\hline
\multicolumn{5}{c}{Wav2vec-based Method \cite{wu+:2020}}                                       \\ \hline
 & \multicolumn{1}{l}{En-Fr} & \multicolumn{1}{c}{$\Delta$} & \multicolumn{1}{l}{En-Ro} & \multicolumn{1}{c}{$\Delta$} \\\cline{2-5}
their baseline$^\dagger$   & 27.8 & \multicolumn{1}{c}{-} & 17.1 & \multicolumn{1}{c}{-} \\
+ wav2vec PT$^\dagger$    & 29.8 & \textcolor{tgreen}{+2.0}                    & 18.2 & \textcolor{tgreen}{+1.1}                  \\
+ vq-wav2vec PT$^\dagger$ & 28.6 & \textcolor{tgreen}{+0.8}                  & 17.4 & \textcolor{tgreen}{+0.3}                  \\ \hline \hline
\multicolumn{5}{c}{MAM-based Method}                                           \\ \hline
our baseline     & 28.9 & \multicolumn{1}{c}{-} & 21.1 & \multicolumn{1}{c}{-} \\
+ MAM w/ English PT        & 31.2 & \textcolor{tgreen}{\textbf{+2.3}}                 & 23.8 & \textcolor{tgreen}{\textbf{+2.7}}                  \\ \hline
\end{tabular}}
\caption{Comparisons between wav2vec-based pre-train method for E2E-ST. Results that are decorated with $^\dagger$ are from \citet{wu+:2020}. Our relative improvements over baseline methods are much larger than wav2vec-based pre-training methods. See more discussions in Sec.~\ref{sec:wav2vec}.}
\label{tb:wav2vec}
\vspace{-5pt}
\end{table}
\vspace{-5pt}
\subsection{Comparisons in Low and Mid-resource Settings}
\label{sec:lowresource}

In Table~\ref{tb:different_resource}, 
we reduce the size of MuST-C from 408 hours to
50 hours and 200 hours to mimic the low and mid-resource language speech translation.

In the scenario when the source language is extremely low-resource 
(no transcribed pre-training and fine-tuning data),
we have ``E2E-ST'' as the baseline.
MAM in both multilingual and acoustic pre-training boosts the performance 
significantly.

When the transcription is only available at the fine-tuning stage (compare with
``+ASR MTL''), MAM pre-trained with Libri-Light achieves similar performance
without using transcription in fine-tuning.

In the cases  when there is no transcription in the fine-tuning stage,
but there exist a large scale annotated pre-training corpus,
MAM still achieves similar performance in 200 hours training setting
without using any transcription.

\begin{table}[th]
\centering
\resizebox{0.9\columnwidth}{!}{
\begin{tabular}{lrrrr}
\hline
\multirow{2}{*}{Models} & \multicolumn{2}{c}{Fr}                             & \multicolumn{2}{c}{Es}                             \\ \cline{2-5}
                        & \multicolumn{1}{c}{50h} & \multicolumn{1}{c}{200h} & \multicolumn{1}{c}{50h} & \multicolumn{1}{c}{200h} \\ \hline
E2E-ST             & 0.52  & 19.83 & 0.4   & 16.13 \\
E2E-ST+ASR MTL     & 8.84  & 25.64 & 7.67  & 20.21 \\
E2E-ST+ASR PT$^*$      & 12.50 & 24.59 & 11.80 & 19.35 \\ \hline \hline
MAM                & 0.6   & 20.54 & 0.4   & 16.85 \\
MAM w/ English PT  & 6.84  & 24.86 & 6.53  & 19.17 \\
MAM w/ acoustic PT & 3.29  & 22.22 & 2.46  & 17.98 \\ \hline
\end{tabular}}
\caption{
   Experimental comparisons with difference training resource.
   $^*$ denotes pretrained with Librispeech corpus.
   See Section~\ref{sec:lowresource} for detailed discussion.}
   \label{tb:different_resource}
\end{table}

\vspace{-5pt}

%% file: related.tex
Text-based BERT-style \cite{BERT,RoBERTa,SpanBERT,ERNIE} frameworks
are extremely popular in recent years due to the remarkable improvement
that they bring to the downstream tasks at fine-tuning stages.
Inspired by techniques mentioned above, MAM also performs 
self-supervised training that masks certain portions randomly 
over the input signals.
But different from BERT-style pre-training,
MAM tries to recover the missing semantic information (e.g., words,
subword units)
and learns the capabilities to restore
the missing speech characteristics and generate the original speech.

SpecAugment~\cite{Park_2019} was originally proposed for ASR as a
data augmentation method by applying a mask over input speech, 
then it is adapted to ST by~\citet{Bahar:2019}.
However, there is no recovering step in SpecAugment, and it 
can not be used as a pre-training framework.

For the self-supervised training in speech domain, 
\citet{Chuang2019,WangWLY020ACL,WangSemantic2020} proposed 
to use forced-alignment to segment speech audio into pieces at
word level and masked some certain words during fine-tuning.
Obviously the forced-alignment based approaches 
rely on the transcriptions of source speech,
and can not be applied to low or zero resource source speech
while MAM will relief the needs of large-scale, 
annotated speech
and translation pairs during pre-training.

\citet{baevski2020wav2vec} proposed 
wav2vec 2.0 pre-training model for ASR task 
which masks the speech input in the latent space and 
pre-trains the model via contrastive task defined 
over a quantization of the latent representations.
In contrast, as the objective of MAM is much simpler and straightforward,
we don't need much extra fine-tuning efforts given an E2E-ST
baseline and massive computational resource.
Furthermore, wav2vec 2.0 is build upon discretized, fix-size,
quantized codebooks, and it is not easy to be adapted to 
arbitrary acoustic signal pre-training.
Lastly, wav2vec 2.0 is designed to have ASR as 
the downstream task, and the 
fine-tuning stage relies on CTC loss \cite{CTC}
which is not straightforward to be adapted in translation task since
translation usually involves with many reordering between
target and source side while
CTC depends on monotonic transition 
function on source side.

%% file: appendix.tex
\setcounter{figure}{0}
\renewcommand{\thefigure}{A\arabic{figure}} 

\setcounter{table}{0}
\renewcommand{\thetable}{A\arabic{table}}

\clearpage
\section*{\Large Appendix}
\appendix

\begin{figure}[h]
\centering
\begin{subfigure}[b]{0.48\textwidth}
   \includegraphics[width=1\linewidth]{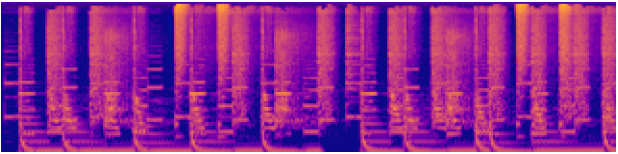}
   \caption{The original musical spectrogram that is mixed with different
   instruments' sound.}
   \label{fig:fmaoriginal}
\end{subfigure}
\begin{subfigure}[b]{0.48\textwidth}
   \includegraphics[width=1\linewidth]{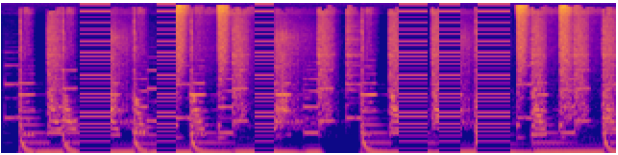}
   \caption{We mask the selected frames (underlined with blue lines) with the same random initialized vector.}
   \label{fig:fmamasked}
\end{subfigure}
\begin{subfigure}[b]{0.48\textwidth}
   \includegraphics[width=1\linewidth]{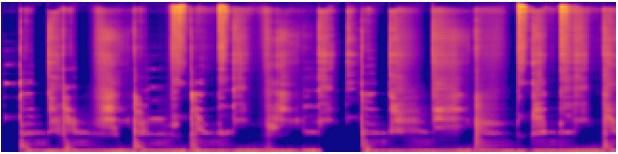}
   \caption{Recovered spectrogram with MAM, pre-trained with Libri-Light corpus.}
   \label{fig:fma2_rec}
\end{subfigure}
\caption{One speech example to showcase the reconstruction ability of pre-trained MAM. Pre-trained MAM with Libri-Light corpus (only human speech data)
can not reconstruct the original music spectrogram accurately since there are
many different musical instruments' sound that is unseen in speech data.}
\label{fig:recfma}
\end{figure}

In the other way around, 
we also try to use Libri-Light pretrained MAM
to recover the corrupted music in Fig.~\ref{fig:recfma}.
MAM that pre-trained with human speech data
does not show good reconstruction in Fig.~\ref{fig:fma2_rec}
since there are many different musical 
instruments' sounds that are unseen in speech data.

